\pdfoutput=1

\documentclass[11pt]{article}

\usepackage[]{naacl2021}
\usepackage{amsmath}
\usepackage{amsfonts}
\usepackage{appendix}
\usepackage{float}
\usepackage{textcomp}

\usepackage{microtype}
\usepackage{todonotes}
\presetkeys{todonotes}{inline}{}
\usepackage{booktabs}

\usepackage{times}
\usepackage{latexsym}

\newcommand{\squad}{\textsc{SQuAD}}

\newcommand{\mppi}{\textsc{MPPI}}

\newcommand{\lstm}{\textsc{LSTM}}

\newcommand{\bert}{\textsc{Bert}}
\newcommand{\xlnet}{\textsc{XLNet}}
\newcommand{\abr}[1]{\textsc{#1}}
\newcommand{\respace}{\vspace*{-.1cm}}

\definecolor{coloranswer}{HTML}{0596FF}
\newcommand*{\mybox}[2]{\tikz[anchor=base,baseline=0pt,rounded corners=0pt, inner sep=0.2mm] \node[fill=#1!60!white] (X) {#2};}

\usepackage[T1]{fontenc}

\usepackage[utf8]{inputenc}

\usepackage{microtype}

\title{On the Transferability of Minimal Prediction Preserving Inputs in Question Answering}
\newcommand*\samethanks[1][\value{footnote}]{\footnotemark[#1]}

\author{Shayne Longpre\thanks{\hspace{2mm}equal contribution} \\
  Apple Inc. \\
  \texttt{slongpre@apple.com} \\\And
  Yi Lu\samethanks \\
  Apple Inc. \\
  \texttt{ylu7@apple.com} \\\And
  Christopher DuBois \\
  Apple Inc. \\
  \texttt{cdubois@apple.com} \\}

\date{}

\begin{document}
\maketitle
\begin{abstract}
Recent work \citep{feng2018pathologies} establishes the presence of short, uninterpretable input fragments that yield high confidence and accuracy in neural models.
We refer to these as Minimal Prediction Preserving Inputs (\mppi{}s).
In the context of question answering, we investigate competing hypotheses for the existence of \mppi{}s, including poor posterior calibration of neural models, lack of pretraining, and ``dataset bias" (where a model learns to attend to spurious, non-generalizable cues in the training data). 
We discover a perplexing invariance of \mppi{}s to random training seed, model architecture, pretraining, and training domain. 
\mppi{}s demonstrate remarkable transferability across domains --- achieving significantly higher performance than comparably short queries.
Additionally, penalizing over-confidence on \mppi{}s fails to improve either generalization or adversarial robustness.
These results suggest the interpretability of \mppi{}s is insufficient to characterize generalization capacity of these models.
We hope this focused investigation encourages more systematic analysis of model behavior outside of the human interpretable distribution of examples.
\end{abstract}

\respace
\section{Introduction}
\respace
\label{sec:introduction}

\citet{feng2018pathologies} establish the presence of shortened input sequences that yield high confidence and accuracy for non-pretrained neural models. 
These Minimal Prediction Preserving Inputs (\mppi{}s) are constructed by iteratively removing the least important word from the query to obtain the shortest sequence for which the model's prediction remains unchanged (example shown in Figure~\ref{fig:example-mppi}).\footnote{For question answering we construct \mppi{}s by only removing words from the query. Modifying the context paragraph is poorly defined in \mppi{} generation as it perturbs the output space, rendering an answer impossible or trivial.} 
Humans are unable to make either confident or accurate predictions on these inputs. 
Follow up work treats strong model performance on such partial-inputs as equivalent with models improperly learning the task \citep{feng2019misleading, kaushik2018much, he2019unlearn}.
Accordingly, we evaluate this proposition in question answering (QA), investigating the properties of \mppi{}s and how their existence relates to ``dataset bias", out-of-domain generalization, and adversarial robustness.

\begin{figure}[t]
	\small
	\tikz\node[fill=white!90!black,inner sep=1pt,rounded corners=0.3cm]{
		\begin{tabular}{lp{0.7\columnwidth}}
			\textbf{\squad{}} \\
			Context & ... 
			The site currently houses three cinemas, including the restored \mybox{coloranswer}{Classic} 
			the United Kingdom's last surviving news cinema still in full-time operation—alongside 
			two new screens ... \\\\
			Original & What's the name of the United Kingdom 's sole remaining news cinema ? \\
			Reduced & news  \\
			Confidence & 0.57 $\to$ 0.51 \\
		\end{tabular}
	}; 
	\caption{A \squad{} dev set example. Given the original
		\emph{Context}, the model makes the same correct prediction (``Classic'') on the \emph{Reduced} question (\mppi{}) as the
		\emph{Original}, with almost the same score. For humans, the reduced
		question, ``news'', is nonsensical.}
	\label{fig:rq_example}
    \vspace*{-.2cm}
    \label{fig:example-mppi}
\end{figure}

First we examine the hypothesis that \mppi{}s are a symptom of poor neural calibration.
\citet{feng2018pathologies} propose we can ``attribute [these neural] pathologies primarily to the lack of accurate uncertainty estimates in neural models.” 
As neural models tend to overfit the log-likelihood objective by predicting low-entropy distributions \citep{guo2017calibration} this can manifest in over-confidence on gibberish examples outside of the training distribution \citep{goodfellow2014explaining}.
We test this hypothesis using pretrained models, shown to have better posterior calibration and out-of-distribution robustness \citep{hendrycks2020pretrained, desai2020calibration}.
Contrary to expectations, we find large-scale pretraining does not produce more human interpretable \mppi{}s.

Second we examine the hypothesis that \mppi{}s are the symptom of ``dataset bias" --- where a flawed annotation procedure results in hidden linguistic cues or ``annotation artifacts" \citep{gururangan-etal-2018-annotation, niven2019probing}.
Models trained on such data distribution can rely on simple heuristics rather than learning the task. 
As such, input fragments or ``partial inputs" are often sufficient for a model to achieve strong performance on flawed datasets.
This explanation has been considered for both Natural Language Inference tasks (the ``hypothesis-only" input for \citet{poliak2018hypothesis, gururangan-etal-2018-annotation}) and for Visual Question Answering (the ``question-only" model for \citet{goyal2017making}).
We expect models which rely on these spurious cues would fail to generalize well to other ``domains" (datasets with different collection and annotation procedures). 
We discover even models trained in different domains perform nearly as well on \mppi{}s as on full inputs, contradicting this hypothesis.
Further, we test their transferability across a number of other factors, including random training seed, model size, and pretraining strategy, and confirm their invariance to each of these.

Third we examine the hypothesis that \mppi{}s inhibit generalization.
This intuition is based on \mppi{}'s poor human interpretability, which could suggest models should not attend to these signals.
To test this hypothesis, we regularize this phenomenon directly to promote more human understandable \mppi{}s, and measure the impact on out-domain generalization and adversarial robustness.
Interestingly, out-domain generalization and robustness on Adversarial \squad{} \citep{jia2017adversarial} vary significantly by domain, with both declining slightly on average due to regularization.

In conjunction, these results suggest \mppi{}s may represent an unique phenomenon from what previous work has observed and analyzed. 
The performance of these inputs is not well explained by domain-specific biases, or posterior over-confidence on out-of-distribution inputs. 
Instead, this behavior may correspond to relevant signals as the impact of their partial mitigation suggests. 
We hope these results encourage researchers to not assume \mppi{}s, or other uninterpretable model behaviour, are dataset artifacts that require mitigation a priori. 
Before presenting mitigation solutions, we propose they follow a more systematic analysis proposed by our actionable framework by (a) rigorously testing the alleged causes of the observed behaviour, (b) confirming the bias does not generalize/transfer, and (c) ensuring the solution provides \textit{consistent} improvements across domains within a task.

\respace
\section{Experimental Methodology}
\respace
\label{sec:exp-meth}

All models trained, including \abr{DrQA} \citep{chen2017reading}, \bert{} \citep{devlin2019bert}, and \xlnet{} \citep{yang2019xlnet}, employ setup and parameter choices from \citet{longpre2019exploration}.\footnote{For \abr{DrQA}, we borrowed the hyper-parameters from hitvoice (\url{https://github.com/hitvoice/DrQA}))}
We generate \mppi{}s by iteratively removing the least important word from the question, while keeping the original prediction unchanged. 
The least important word is given as that for which the model's confidence in its prediction remains highest in its absence.\footnote{Details of model training and examples of \mppi{} generation are described in Appendix A.}

\begin{table}[t]
\centering
\resizebox{\linewidth}{!}{\begin{tabular}{lccc}
{Dataset} & \textsc{Original} & \textsc{Bert-B} & \textsc{XLNet-L} \\
\toprule
\abr{SQuAD} \citep{rajpurkar2016squad} & 11.54 & 2.32 & 2.65 \\
\abr{HotpotQA} \citep{yang2018hotpotqa}& 18.96& 2.07 & 2.55 \\
\abr{NewsQA} \citep{trischler2016newsqa}& 7.59 & 2.08 & 1.80 \\
\abr{NaturalQ} \citep{kwiatkowski2019natural}& 9.17& 1.22 & 1.26 \\
\abr{TriviaQA} \citep{joshi2017triviaqa}& 15.68 & 2.33 & 1.80 \\
\abr{SearchQA} \citep{dunn2017searchqa}& 17.43 & 1.81 & 1.05 \\
\bottomrule
\end{tabular}}
\caption{\label{dataset-mppi-length}
Number of \mppi{} query tokens, for different datasets and models.
}
\vspace*{-.4cm}
\end{table}

To examine how \mppi{}s transfer across Question Answering domains we employ 6 diverse QA training sets and 12 evaluation sets.\footnote{Refer to Appendix A.3 for details, or the MRQA 2019 workshop: \url{https://mrqa.github.io/shared}. \citet{fisch2019mrqa} normalized these datasets into purely answerable, extractive format.}
The datasets were selected for annotation variety, differing on: question type, document source, annotation instructions, whether the question was collected independently of the passage, and skills required to answer the question. 
This set represents a realistic spectrum of domains for evaluating generalization. 

We set aside $2k$ examples from each domain's validation sets in order to generate \mppi{}s for model evaluation.
For each experiment we also generate a set of randomly shortened queries to compare against the \mppi{}s --- we refer to this as the ``Random \mppi{}" baseline.
For each of the original examples, we generate this baseline by randomly removing words until the length matches that of the corresponding \mppi{}.

\respace
\section{Experiments}
\respace
\label{sec:experiments}

\subsection{Invariance of \mppi{}s}
\label{sec:invariance}

\citet{feng2018pathologies} establish the ``human-insufficiency" property of \mppi{}s for non-pretrained, \lstm{} and attention-based models, including \abr{DrQA}, and \abr{BiMPM} \citep{wang2017bilateral}. 
We extend this investigation for modern, pretrained Transformers, and assess the ``invariance" of \mppi{}s: measuring whether they are random, or are affected by model architecture, pretraining strategy, or training dataset (domain).

\begin{table}
	\small
	\resizebox{\linewidth}{!}{\begin{tabular}{lccc}
			{} & \abr{DrQA} & \textsc{Bert-B} & \textsc{XLNet-L} \\
			\toprule{}
			\textsc{Bert-B} & 32.1 / 9.9 & - / - & 29.8 / 9.9\\
			\textsc{ XLNet-L} & 26.0 / 7.2 & 29.8 / 9.9  & - / - \\
			\midrule{}
			\abr{Random} & 13.0 / 1.8 & 12.6 / 0.9 & 14.2 / 1.3 \\
			\bottomrule
	\end{tabular}}
	\caption{\label{model-mppi-similarity}
		The mean similarity, measured in Jaccard Similarity / Exact Match (\%), between the \mppi{}s from different model types and the random baseline. 
	}
	\vspace*{-.4cm}
\end{table}

In subsequent experiments we compare sets of \mppi{}s using the mean Exact Match or Generalized Jaccard Similarity (GJS), a variant of Jaccard Similarity, which accounts for the possibility of repeated tokens in either of the sequences being compared.
Generalized Jaccard Similarity is defined between two token sequences $X$ and $Y$ in Equation \ref{eq:gjs}. 
Here, $n$ is the index of every element that appears in $X\cup Y$.
	
\begin{equation}\label{eq:gjs}
GJS(X, Y) = \frac{\sum_{i=1}^{n}min(X_i,Y_i)}{\sum_{i=1}^{n}max(X_i,Y_i)}
\end{equation}

We will refer to this as ``Jaccard Similarity" for simplicity.

\respace
\subsubsection{Random Seed}
\respace

First, we investigate whether \mppi{}s are ``random", or influenced by weight initialization and training data order. 
Measuring the Jaccard Similarity between \mppi{} sequences produced by models with different training seeds we find $\text{JS}_{\text{MPPI}}=57.1\%\pm1.2$, as compared to $\text{JS}_{R}=13.8\%\pm0.8$ on the Random \mppi{} baseline.
This suggests \mppi{}s are not simply the side-effect of randomness in the training procedure.

\respace
\subsubsection{Pretraining and Architecture}
\respace

One hypothesis is that traditional \lstm{}-based models, such as \abr{DrQA}, do not have sufficient pretraining or ``world knowledge" to rely on the entire sequence, and overfit to subsets of the input. If this were the primary source of \mppi{}s, we might expect models that are better calibrated and more robust to out-of-distribution examples to have longer and more interpretable \mppi{}s. 
Accordingly, we test this hypothesis with large pretrained transformers, which recent work demonstrates have better posterior calibration and robustness to out-of-distribution inputs. 

Specifically, \citet{desai2020calibration} examine 3 separate NLP tasks, using ``challenging out-of-domain settings, where models face more examples they should be uncertain about", and find that ``when used out-of-the-box, pretrained models are calibrated in-domain, and compared to baselines, their calibration error out-of-domain can be as much as 3.5× lower". 
Similarly \citet{hendrycks2020pretrained} systematically show ``Pretrained transformers are also more effective at detecting anomalous or [out-of-distribution] examples".
These findings suggest pretrained transformers should produce more interpretable \mppi{}s than non-pretrained models.

However, in Table~\ref{dataset-mppi-length} we show \mppi{}s remain incomprehensibly short for all 6 domains and even for pretrained transformer models 
(\abr{DrQA} produces \mppi{}s on \squad{} of mean length $2.04$).
In Table~\ref{model-mppi-similarity} we show \mppi{}s produced by different model architectures and pretraining strategies are similar, significantly exceeding the Jaccard Similarity of the Random \mppi{} baseline ($\text{JS}_{\text{R}}=13.8\%$).
This would not be problematic if pretrained models produced lower confidences for \mppi{}s than the original examples (demonstrating some form of calibration).
However, we find the opposite is true.
Taking SQuAD for instance we see in 85\% of cases the BERT model is more confident on the \mppi{} than the original example.

Lastly, we verify with manual grading tasks that the \mppi{}s for BERT and XLNet are no more interpretable to humans than DrQA’s \mppi{}s, as shown in Table~\ref{human-f1}.
This suggests that short, uninterpretable \mppi{}s are ubiquitous in modern neural question answering models and unmitigated by large scale pretraining, or improved out-of-distribution robustness.

\begin{table}[t]
\resizebox{\linewidth}{!}{\begin{tabular}{lcccc}
{\textbf{Train Dataset}} & \multicolumn{4}{c}{\textbf{Reduction Dataset}} \\
\toprule{}
{} & \abr{SQuAD} & \abr{HotPotQA} & \abr{NewsQA} & \abr{NatrualQ} \\
\midrule{}
\abr{SQuAD} & - (-) & 31.4 (8.8) & 41.0 (21.6) & 29.2 (12.5)  \\
\abr{HotPotQA} & 39.7 (12.8) & - (-) & 39.6 (18.8) & 33.8 (13.5)  \\
\abr{NewsQA} & 41.1 (13.0) & 31.6 (7.2) & - (-) & 35.2 (12.5)  \\
\abr{NatrualQ} & 37.5 (12.7) & 28.7 (7.1) & 40.2 (17.9) & - (-)  \\
\midrule
{Average} & 39.4 (12.8) & 30.6 (7.7) & 40.3 (19.4) & 32.7 (12.8) \\
\bottomrule
\end{tabular}}
\caption{\label{dataset-mppi-similarity}
The Jaccard Similarity (\%) between \abr{BERT} generated \mppi{}s, across domains. 
In parentheses are the Jaccard Similarity scores between the Random \mppi{} baseline and Train Dataset \mppi{}s.}
\vspace*{-.4cm}
\end{table}

\respace
\subsubsection{Cross-Domain Similarity}
\respace

Next, we investigate the extent to which \mppi{}s are domain-specific.
We do this by measuring their similarity when produced by models trained in different domains.
If \mppi{}s are the product of bias in the training data, such as annotation artifacts, we would expect them to be relatively domain specific, as different datasets carry different biases.
In Table~\ref{dataset-mppi-similarity} a model trained from each domain (Train Dataset) generates \mppi{}s for each other domain (Reduction Dataset). 
For each Reduction Dataset, we measure the mean Jaccard Similarity between \mppi{}s produced by the Train Dataset model and \mppi{}s produced by the Reduction Dataset (in-domain) model.
In parentheses we show the mean Jaccard Similarity between the Random \mppi{}s and the Train Dataset \mppi{}s.
In all cases, \mppi{}s demonstrate higher similarity than the random baseline, indicating that they are not domain specific. 

\respace
\subsection{Cross-Domain Transferability of \mppi{}s}
\respace
\label{sec:xdomain-gen}

Even when models generate different \mppi{}s, they may still transfer to the other domain.
We would like to measure \mppi{} transferability, independent of their similarity between models.
If QA models perform well on \mppi{}s generated from a range of domains then this would suggest they are not a product of bias in the training data.
Instead, they may retain information important to question answering, rather than annotation artifacts.
To better measure the extent of \mppi{} transferability, we (a) train one model on SQuAD (Train Dataset), and another on NewsQA (Reduction Dataset), (b) use the NewsQA-model to generate $2k$ \mppi{}s on the NewsQA evaluation set, and (c) measure the F1 performance of the SQuAD-model evaluated on both the original NewsQA evaluation set and the \mppi{} queries as generated in part (b). 

\begin{figure}[h]
  \includegraphics[width=1\linewidth]{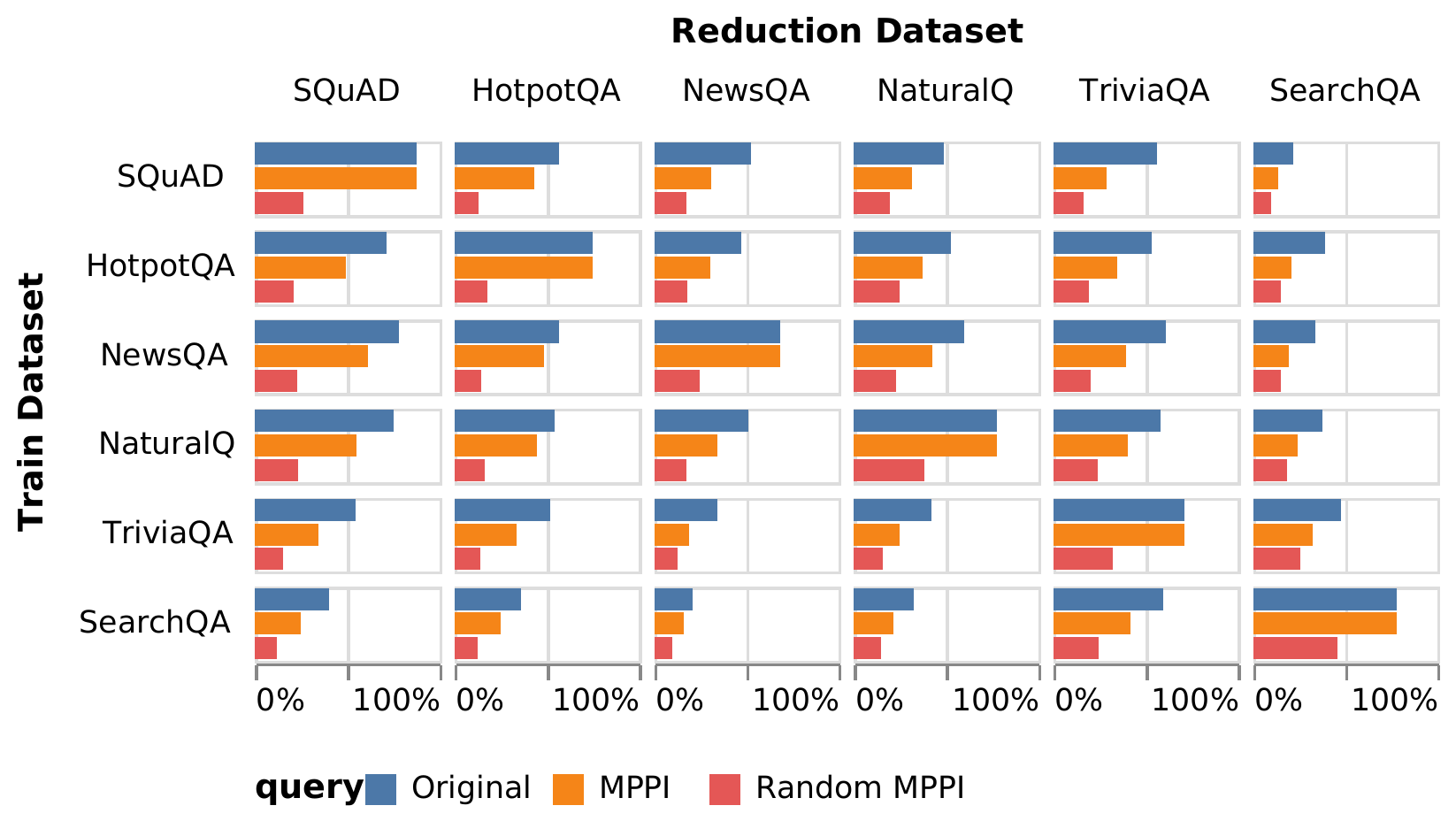}
  \caption{Cross-Domain Transferability: \abr{BERT} question answering performance (F1) with different training sets (y-axis), and $2k$ evaluation sets (x-axis). Bars are colored by input type. On average, \mppi{} queries close the gap between models' performance on Random \mppi{}s and original queries by $46.6\%$.}
  \label{fig:cross-domain-generalization-combined-v0}
\end{figure}

Figure~\ref{fig:cross-domain-generalization-combined-v0} shows performance on out-domain \mppi{}s are $46.6\%$ closer to original performance than on Random \mppi{}s.
This evidence suggests \mppi{}s are highly transferable across domains.
Consequently, \mppi{}s may relate to generalization, despite their poor human interpretability.

\respace
\respace
\subsection{Human-Sufficient \mppi{}s do not Improve Generalization}
\respace
\label{sec:entropy-pen-generalization}

Even though \mppi{}s are highly transferable between domains, their presence may be associated with poor generalization.
To evaluate this possibility, we examine whether the penalization of \mppi{}s improves generalization, or adversarial robustness. 
While penalizing over-confidence on \mppi{}s has been shown to maintain equivalent in-domain performance, and yield subsequently longer and more human interpretable \mppi{} queries \citep{feng2018pathologies}, its impact on generalization or robustness has not yet been examined.

\begin{table}[t]
\centering
\resizebox{\linewidth}{!}{\begin{tabular}{lccccc}
\toprule
Train Dataset & \multicolumn{5}{c}{F1 Score (\%)} \\
\midrule
{} & $\triangle$ID & OD & $\triangle$OD & AR & $\triangle$AR \\
\midrule
\abr{SQuAD} & -0.8 & 52.9 & -1.5 $\pm$ 2.3 & 72.1 & +3.1 \\
\abr{HotpotQA} & +0.6 & 48.5 & -0.6 $\pm$ 1.2 & 45.5 & +1.0 \\
\abr{NewsQA} & -0.9 & 53.0 & -0.9 $\pm$ 0.6 & 62.9 & -1.8 \\
\abr{NaturalQ} & +0.9 & 51.6 & -2.9 $\pm$ 3.5 & 54.9 & -0.9 \\
\abr{TriviaQA} & -0.6 & 42.3 & -4.1 $\pm$ 2.8 & 38.9 & -1.1 \\
\abr{SearchQA} & -0.5 & 38.0 & -5.9 $\pm$ 2.9 & 32.3 & -4.0 \\
\midrule
\abr{Overall Avg} & -0.2 & 47.7 & -2.7 $\pm$ 1.1 & 51.1 & -0.6 \\
\bottomrule
\end{tabular}}
\caption{\label{reg-vs-pathos-f1}
The impact of \mppi{} regularization on in-domain (ID) performance, macro-average out-domain (OD) generalization over 12 evaluation datasets, and adversarial robustness (AR) on Adversarial \squad{}.
$\triangle$X = F1 of \mppi{} regularized model minus F1 of regular model on target X (any of ID, OD, or AR).}
\vspace*{-.4cm}
\end{table}

We employ a simplified version of the \mppi{} penalization used by \citet{feng2018pathologies}, training a model with equal quantities of regular and \mppi{} examples --- maintaining normal QA loss terms for the regular examples, and applying an entropy penalty to \mppi{} examples.\footnote{See Appendix section A.4 for details.} 
When penalizing over-confidence on \mppi{}s, we confirm the new \mppi{} length is significantly longer (Appendix sections B), and more human interpretable (Table~\ref{human-f1}). 

In Table~\ref{reg-vs-pathos-f1} we show the difference in F1 scores ($\triangle$) between the regularized and original models.
Results demonstrate that in-domain F1 (ID), macro-average out-domain F1 over 12 datasets (OD), and adversarial robustness F1 on Adversarial \squad{} (AR) all decline slightly on average with \mppi{} regularization --- by $0.2\%$, $2.7\%$, and $0.6\%$ respectively.
These results suggest a model's ability to make predictions on \mppi{}s is not strongly correlated with either generalization or robustness across 13 total QA datasets. 
However, the relative stability of in-domain performance as compared to out-domain performance suggests mitigating \mppi{}s is more harmful when crossing domain boundaries.

\begin{table}[t]
	\small
	\centering
	\begin{tabular}{lr}
		& \textsc{Human F1 (EM)} \\
		\toprule
		\textsc{Original Query} &  91.2 (82.3) \\
		\abr{DrQA$^\dagger$} MPPI & - (31.7) \\
		\abr{BERT-B} MPPI &  41.6 (32.0) \\
		\abr{XLNet-L} MPPI & 37.6 (26.0) \\
		\midrule
		\abr{BERT-B*} MPPI & 60.7 (43.5) \\
		\midrule
		\abr{Random} MPPI & 26.5 (17.0) \\
		\bottomrule
	\end{tabular}
	\caption{\label{human-f1}
		The mean human performance (in F1 and Exact Match over 100 examples) on different variants of \mppi{}s for \squad{}. \\
		$^\dagger$ Human performance cited from \citet{feng2018pathologies} \\
		$^*$ Indicates a model trained with \mppi{} regularization.
	}
\end{table}

Certain train datasets exhibit greater sensitivity to \mppi{} regularization than others.
For instance SearchQA is drastically affected in all measures, HotpotQA hardly at all, and SQuAD actually improves by $3.1\%$ in adversarial robustness.
Additionally, Table~\ref{reg-vs-pathos-f1} shows the $95\%$ confidence intervals for out-domain generalization are often as large as the mean change in performance.
Empirically, this demonstrates the effect of \mppi{} regularization is not consistent, having both positive and negative impacts on performance, depending on which of the 12 out-domain datasets is in question.\footnote{See Figure 9 in Appendix A.4 for details.}

\respace
\respace
\section{Discussion}
\respace
\label{sec:discussion}

In \squad{}, the most common \mppi{} is the empty string (40\%). 
Among non-empty strings, the most common \mppi{} tokens are: ``what", ``?", ``who", ``how", ``when". 
Despite the pattern of interrogative words, these tokens are already among the most frequent in \squad{} questions, so it’s challenging to measure the unique information they convey.

A more direct approach to understand the informative signal of \mppi{}s is to measure their ``human insufficiency" property directly.
We conduct a grading task, comparing human ability to answer real, \mppi{}, and random \mppi{} queries.
Table~\ref{human-f1} shows that humans could only correctly answer BERT and XLNet \mppi{}s slightly more often than random \mppi{}s (32\% and 26\% exact match compared to 17\%), but could answer 43.5\% of \mppi{}s produced by \mppi{}-regularized BERT.
Although this confirms \mppi{}-regularization partially resolves over-confident behaviour for these human non-interpretable inputs, we've observed the resulting model fares slightly worse in domain generalization and robustness.

We find no evidence that \mppi{}s are explained by poorly calibrated neural models, lack of pretraining knowledge, or dataset-specific bias. 
Alternatively they may relate, at least in part, to useful and transferable signals. 
While practitioners, especially in model debiasing tasks, have focused on human understandable and generalizable features, this work would encourage them to also consider the presence of generalizable features which are not human interpretable.
This observation closely relates to prior work in computer vision suggesting human uninterpretable, adversarial examples can be the result of ``features", not ``bugs", in which \citet{ilyas2019adversarial} observe ``a misalignment between the (human-specified) notion of robustness and the inherent geometry of the data."
We hope this work provides a framework to rigorously evaluate the impact of bias mitigation methods on robustness and generalization, and encourages ML practitioners to examine assumptions regarding unexpected model behaviour on out-of-distribution inputs.

\respace
\respace
\section{Conclusion}
\respace
\label{sec:conclusion}

We empirically verify the surprising invariance of \mppi{}s to random seed, model architecture, and pretraining, as well as their wide transferability across domains. 
These results suggest that \mppi{}s may not be best explained by poorly calibrated neural estimates of confidence or dataset-specific bias. 
Examining their relationship to generalization and adversarial robustness, we highlight the ability to maintain in-domain performance but significantly alter out-domain performance and robustness.
We hope our results encourage a more systematic analysis of hypotheses regarding model behavior outside the human interpretable distribution of examples.

\respace
\respace
\section{Acknowledgments} 
\respace
We would like to acknowledge Eric Wallace, Shi Feng, Jordan Boyd-Graber, Christopher Clark, Drew Frank, Kanit Wongsuphasawat, Ni Lao, and Charlie Maalouf for their guiding insights and helpful discussion.

\clearpage

\bibliography{anthology,custom}
\bibliographystyle{acl_natbib}

\clearpage
\appendix
\appendixpage

\label{sec:appendix}

\section{Reproducibility}
\subsection{Question Answering Models}
For reproducibility, we share our hyper-parameter selection in Table~\ref{appendix-hyperparams}. 
We borrow our hyper-parameters from \citet{longpre2019exploration} for training all Question Answering (QA) models. 
Their parameters are tuned for the same datasets in the MRQA Shared Task. 
We found these choices to provide stable and strong results across all datasets.

Our BERT and XLNet question answering modules build upon the standard PyTorch \citep{NEURIPS2019_9015} implementations from HuggingFace, and are trained on $8$ NVIDIA Tesla V100 GPUs.\footnote{\url{https://github.com/huggingface/transformers}}
For \abr{DrQA}, by \citet{chen2017reading}, we borrowed the implementation and hyper-parameters from hitvoice (\url{https://github.com/hitvoice/DrQA}) and train on $1$ NVIDIA Tesla V100 GPU.\footnote{We used the open source version available at \url{https://github.com/hitvoice/drqa}.} 

\begin{table}[htb]
	\small
	\centering
	\begin{tabular}{ll}
		\toprule
		\textbf{Model Parameter} & \textbf{Value} \\
		\midrule
		\textbf{\abr{DrQA}} & {} \\
		\midrule
		Model Size (\# params) & $31.5M$ \\
		Avg. Train Time & $10h\ 30m$ \\
		\midrule
		Learning Rate & 0.1 \\
		Optimizer & Adamax \\
		Num Epochs & 35 \\
		Batch size & 32 \\
		Dropout & 0.4 \\
		Hidden size & 128 \\
		\midrule
		\textbf{BERT-BASE} & {} \\
		\midrule
		Model Size (\# params) & $108.3M$ \\
		Avg. Train Time & $2h\ 20m$ \\
		\midrule
		Learning Rate & $5e-5$ \\
		Optimizer & Adam \\
		Num Epochs & 2 \\
		Batch Size & 25 \\
		Gradient Accumulation & 1 \\
		Dropout & $0.1$ \\
		Lower Case & False \\
		Max Query Length & 64 \\
		Max Sequence Length & 512 \\
		\midrule
		\textbf{XLNet-LARGE} & {} \\
		\midrule
		Model Size (\# params) & $364.5M$ \\
		Avg. Train Time & $4h\ 45m$ \\
		\midrule
		Learning Rate & $2e-5$ \\
		Optimizer & Adam \\
		Num Epochs & 2 \\
		Batch Size & 6 \\
		Gradient Accumulation & 3 \\
		Dropout & $0.1$ \\
		Lower Case & False \\
		Max Query Length & 64 \\
		Max Sequence Length & 512 \\
		\bottomrule
	\end{tabular}
	\caption{\label{appendix-hyperparams}
		Hyperparameter selection for each model type.
	}
\vspace*{-.5cm}
\end{table}

\subsection{Dataset}

We employ 6 diverse QA training sets and 12 evaluation sets from the MRQA 2019 workshop (\url{https://github.com/mrqa/MRQA-Shared-Task-2019}) \citep{fisch2019mrqa}. 
These datasets have been normalized into purely extractive format and all questions are answerable. 
The 6 training datasets are SQuAD \citep{rajpurkar2016squad}, NewsQA \citep{trischler2016newsqa}, TriviaQA \citep{joshi2017triviaqa}, SearchQA \citep{dunn2017searchqa}, HotpotQA \citep{yang2018hotpotqa}, and Natural Questions \citep{kwiatkowski2019natural}. 
Six other evaluation datasets are included: BioASQ \citep{tsatsaronis2012bioasq}, DROP \citep{dua2019drop}, DuoRC \citep{saha2018duorc}, RACE \citep{lai2017race}, RelationsExtraction \citep{levy2017zero} , and TextbookQA \citep{kembhavi2017you}. 
Table~\ref{dataset-stat} shows their statistics.

\begin{table*}
\footnotesize
\centering
\tabcolsep=0.25cm
\begin{tabular}{lllrrrr@{}}
\textbf{Dataset}                               & \textbf{Question (Q)} & \textbf{Context (C)} & \textbf{{Avg. Q Len}} & \textbf{Avg. C Len}  & \textbf{Train} & \textbf{Dev} \\ 
\toprule
SQuAD & Crowdsourced      & Wikipedia  & 11  &  137  & 86,588         & 10,507       \\
NewsQA & Crowdsourced      & News articles    & 8 & 599  & 74,160         & 4,212        \\
TriviaQA & Trivia            & Web snippets   & 16 & 784 & 61,688         & 7,785        \\
SearchQA & Jeopardy          & Web snippets & 17 & 749 & 117,384        & 16,980       \\
HotpotQA & Crowdsourced      & Wikipedia    &  22 & 232 & 72,928         & 5,904\\
Natural Questions & Search logs       & Wikipedia     & 9  & 153 & 104,071        & 12,836\\ 
\midrule
BioASQ & Domain experts    & Science articles & 11 & 248  & -              & 1,504 \\
DROP & Crowdsourced      & Wikipedia &   11  &  243  & -              & 1,503 \\
DuoRC & Crowdsourced      & Movie plots   & 9 & 681 & -              & 1,501\\
RACE & Domain experts          & Examinations   & 12 & 349 & -              & 674          \\
RelationExtraction   & Synthetic         & Wikipedia    & 9 &  30  & -              & 2,948      \\
TextbookQA & Domain experts    & Textbook & 11   &   657  & - & 1,503\\ 
\bottomrule
\end{tabular}
\caption{Statistics about datasets used: The first block presents six domains used for training, the second block presents six additional domains used for model evaluation and generating \mppi{}s.}
\label{dataset-stat}
\vspace*{-.2cm}
\end{table*}

We use the hyperparameters described in Table~\ref{appendix-hyperparams} for training on each dataset.
We use all the training data provided for each by \textsc{MRQA}.

\subsection{Generating \mppi{}s}

The process for generating \mppi{}s closely follows the procedure described by \citet{feng2018pathologies}. 
We operate with a beam size of $k=3$, finding that larger beam sizes exhibit diminishing returns, and rarely produce different results. 
The procedure involves iteratively removing the token which is ``least important" to the model.
The least important token is defined as the one that when removed provides the smallest decrease in confidence in the originally predicted span.
Note that in some cases confidence in the originally predicted span can even increase with the removal of a token.
In any case, the least important token is always designated by the lowest confidence in the original prediction.
The stop condition is when removing \textbf{any} additional token would change the model's prediction. 

Note that we follow previous work in only removing words from the query in extractive question answering. 
The reason for this is the \mppi{} can be poorly defined when context tokens are removed.
Since the output predictions are over the context tokens for extractive question answering, its possible to warp the answer space, or remove the answer altogether.
Additionally, if we do not permit any alterations to the original prediction tokens, then there exists a trivial solution: remove all tokens except for the predicted answer.
In this case an extractive question answering model is forced to predict that answer, with no alternative options.
Consequently, \mppi{}s that allow modifications to the context, or output space, can be poorly defined.
Since in question answering the query is an essential input to provide confident answers, we believe this is the most reasonable setup for the task.

\begin{equation}\label{eq:1}
p_{ij} = \max(\text{softmax}(S_i + E_j))
\end{equation}

For completeness, we describe our method of computing span confidence for question answering, given that there are many variations. 
Let $S \in \mathbb{R}^{N}$ be the vector of start logits and $E \in \mathbb{R}^{N}$ be the vector of end logits, both of sequence length $N$. 
For every combination of $i,\ j \in [0,N]$ where $j \ge i \le min(j+C,\ N)$, and $C=30$ is the maximum answer span length, we compute the confidence for that span of answer text as the sum of their respective logits $S_i + E_j$. 
The final confidence probability $p_ij$ for a given span is as shown in Equation~\ref{eq:1}.

The model, on the other hand, can still make the same prediction as it did on the full input, and with a similar degree of confidence. 

\subsection{Regularizing \mppi{}s}

There are a couple differences between the \mppi{} entropy-regularization strategy employed in this work and in \citet{feng2018pathologies}. 
While \citet{feng2018pathologies} fine-tune an a model already trained for the question answering task, we regularize \mppi{}s in the initial fine-tuning stage (starting from BERT and XLNet's pre-trained weights). 
Secondly, they alternate updates between two optimizers, one batch of maximum likelihood, two for \mppi{} entropy maximization, whereas we use the same optimizer and shuffle together equal numbers of \mppi{} and regular inputs. 
We find our method (without rigorous comparison) to be slightly more effective on BERT at mitigating the \mppi{} phenomenon (measured by subsequent \mppi{} length). 
We suspect, if there is an advantage, it is due to the regularization beginning with the start of fine-tuning, rather than on a subsequent stage of fine-tuning.

\begin{equation} \label{eq:reg-ent}
\mathcal{L_{MPPI}}=C-\lambda\sum_{\tilde{x}\in \tilde{\mathcal{X}}}
\mathbb{H}\left(f(y | \tilde{x})\right)
\end{equation}

\begin{equation} \label{eq:final-loss}
\mathcal{L}=\mathcal{L_{QA}}+\mathcal{L_{MPPI}}
\end{equation}

For completeness, we provide our entropy regularization loss term in Equation~\ref{eq:reg-ent}. 
Let $\tilde{\mathcal{X}}$ denote the set of inputs that have been reduced to their \mppi{}, $\mathbb{H}\left(\cdot\right)$ denote the entropy and $f(y | x)$ denote the predicted confidence for $y$ given $x$. 
We then represent the loss term for \mppi{}s as $\mathcal{L_{MPPI}}$, where the constant $C=10$ is chosen such that maximizing the entropy will minimize the loss. 
We use $\lambda=0.1$ as the most effective choice in our limited set of trials. 
The full loss term, for all inter-mixed regular question answering, and \mppi{} examples is the sum of standard QA loss $\mathcal{L_{QA}}$, and the \mppi{} loss term $\mathcal{L_{MPPI}}$, as shown in Equation~\ref{eq:final-loss}.

\begin{figure*}
	\centering
	\includegraphics[width=\linewidth]{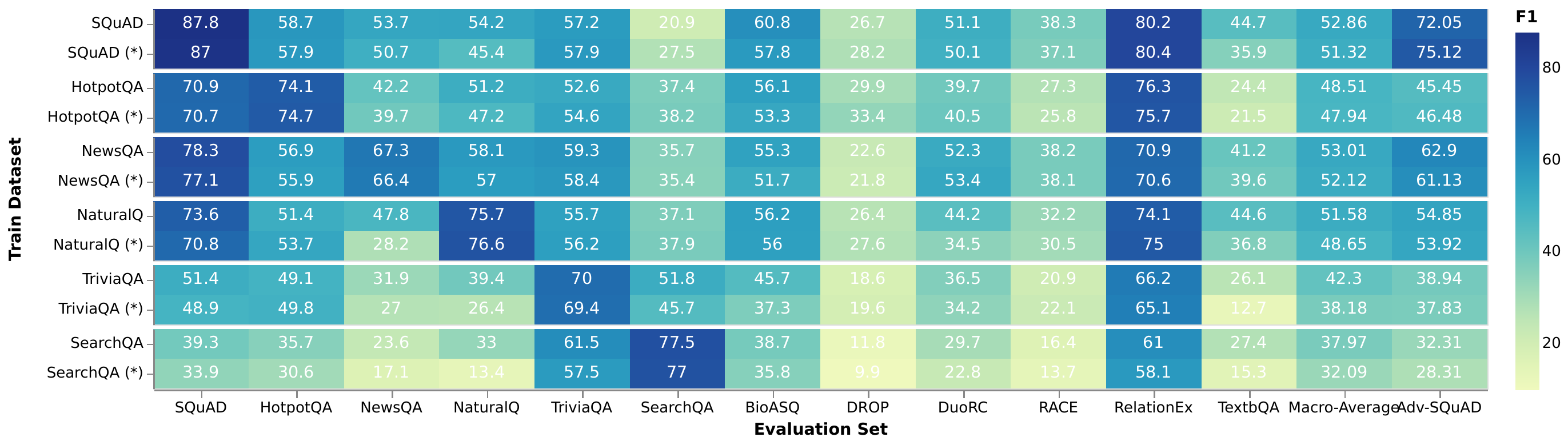}
	\caption{The generalization and robustness of \abr{BERT} models evaluated on 12 datasets, as well as Adversarial SQuAD. The ``(*)" indicates \mppi{}-regularization during training.}
	\label{fig:reg-vs-pathos-f1-appendix}
	\vspace*{-.4cm}
  \end{figure*}

In Figure~\ref{fig:reg-vs-pathos-f1-appendix} we display the full comparison between the performances of the \mppi{} regularized models and the regular models on 13 QA datasets, including Adversarial SQuAD \citep{jia2017adversarial}.

\section{How do \mppi{} Lengths Compare?}

\begin{figure}[H]
	\includegraphics[width=1\linewidth]{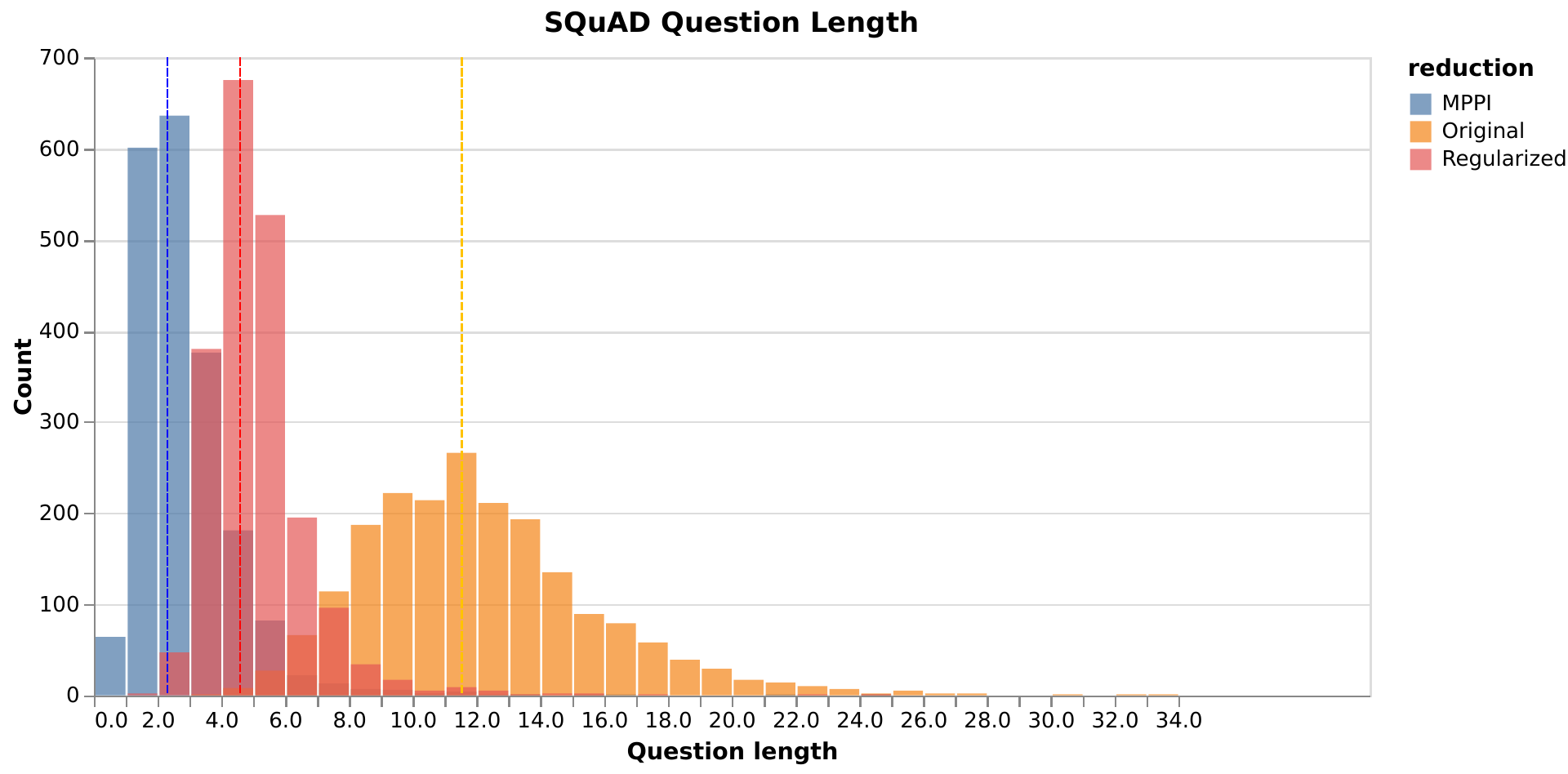}
	\caption{\abr{SQuAD} question length generated by different \mppi{} reduction methods}
	\label{fig:squad-lens}
	\vspace*{-.4cm}
\end{figure}

\begin{figure}[H]
	\includegraphics[width=1\linewidth]{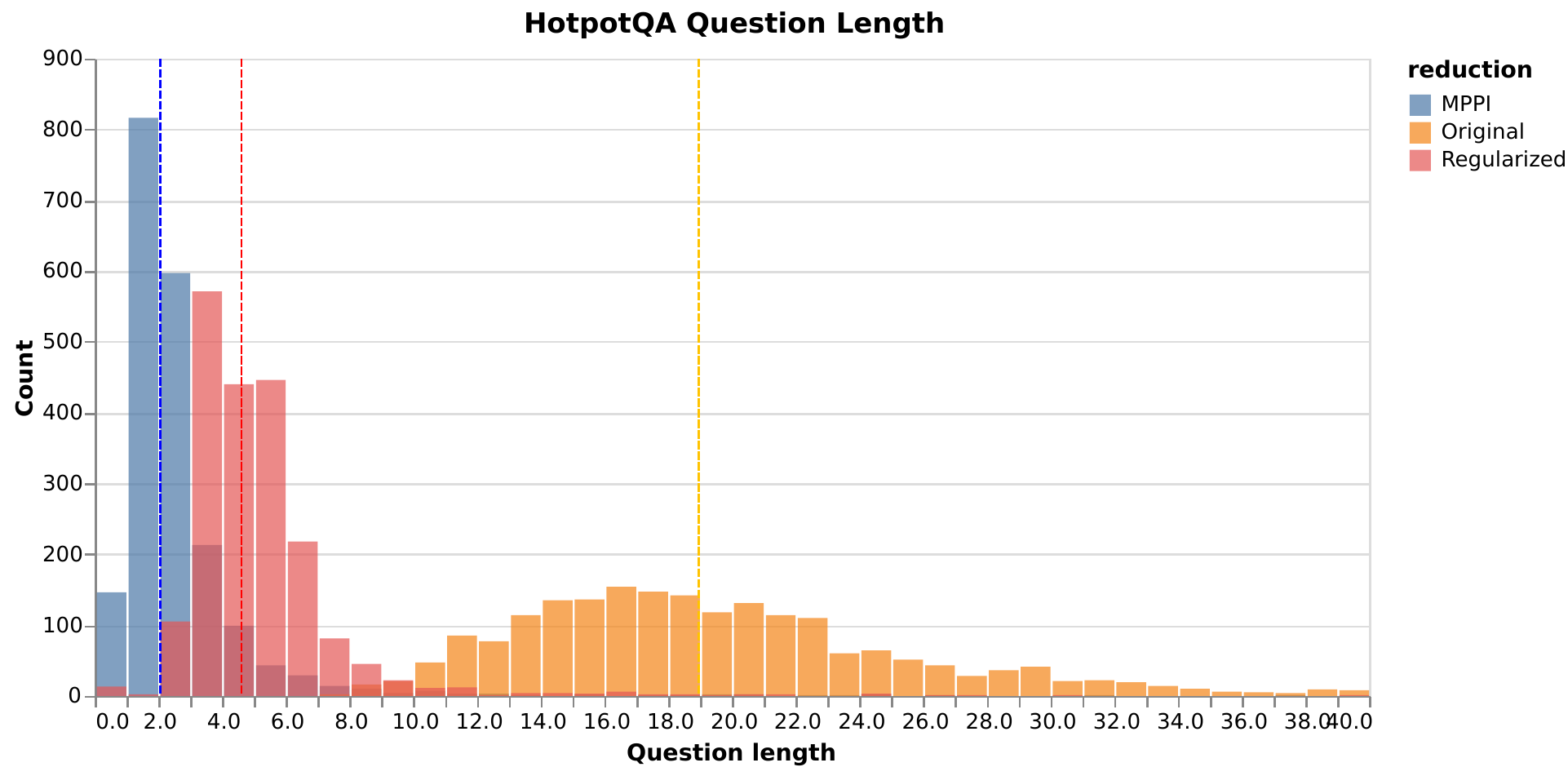}
	\caption{\abr{HotpotQA} question length generated by different \mppi{} reduction methods}
	\label{fig:hotpot-lens}
	\vspace*{-.4cm}
\end{figure}

\begin{figure}[H]
	\includegraphics[width=1\linewidth]{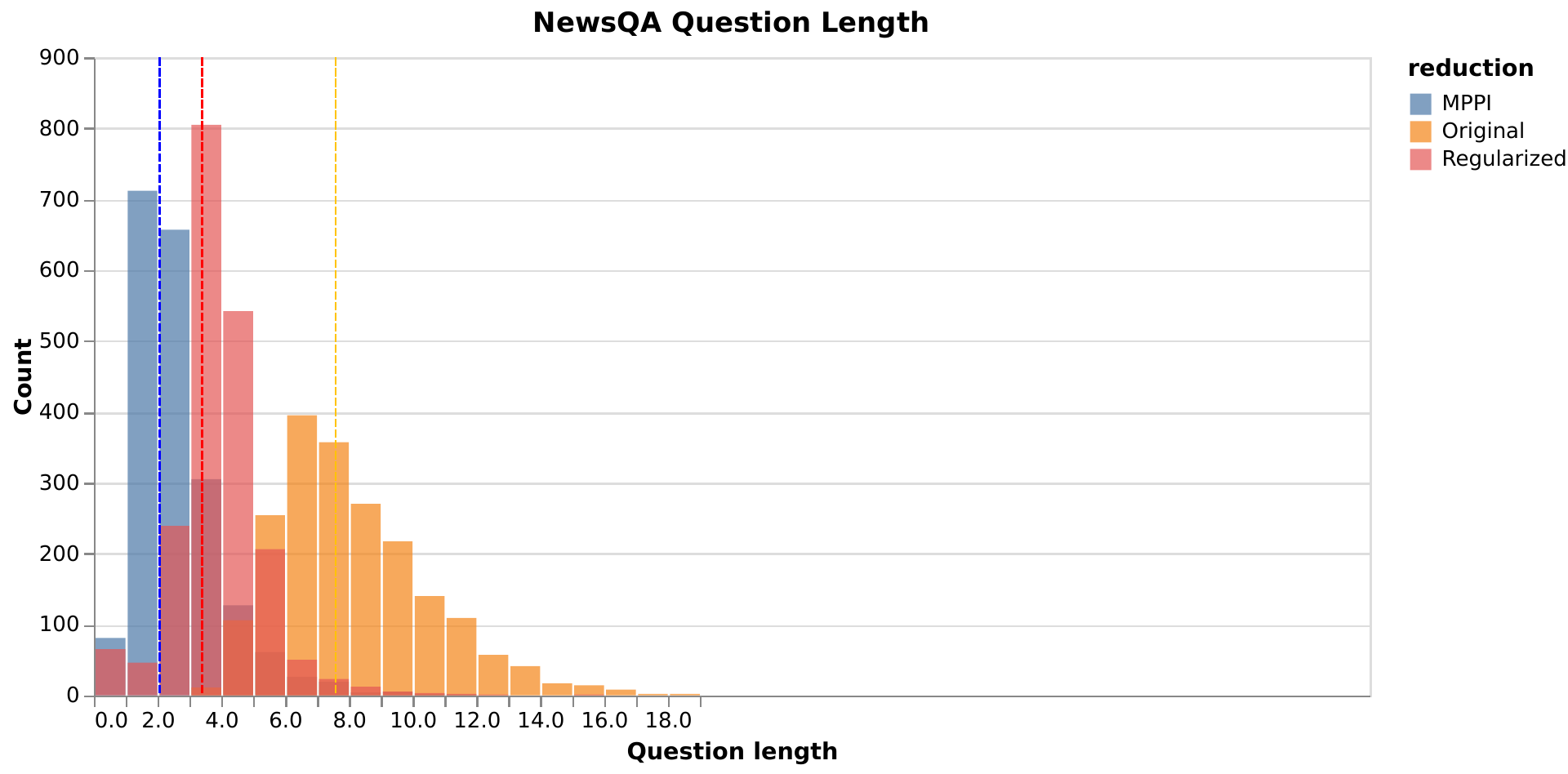}
	\caption{\abr{NewsQA} question length generated by different \mppi{} reduction methods}
	\label{fig:news-lens}
	\vspace*{-.4cm}
\end{figure}

\begin{figure}[H]
	\includegraphics[width=1\linewidth]{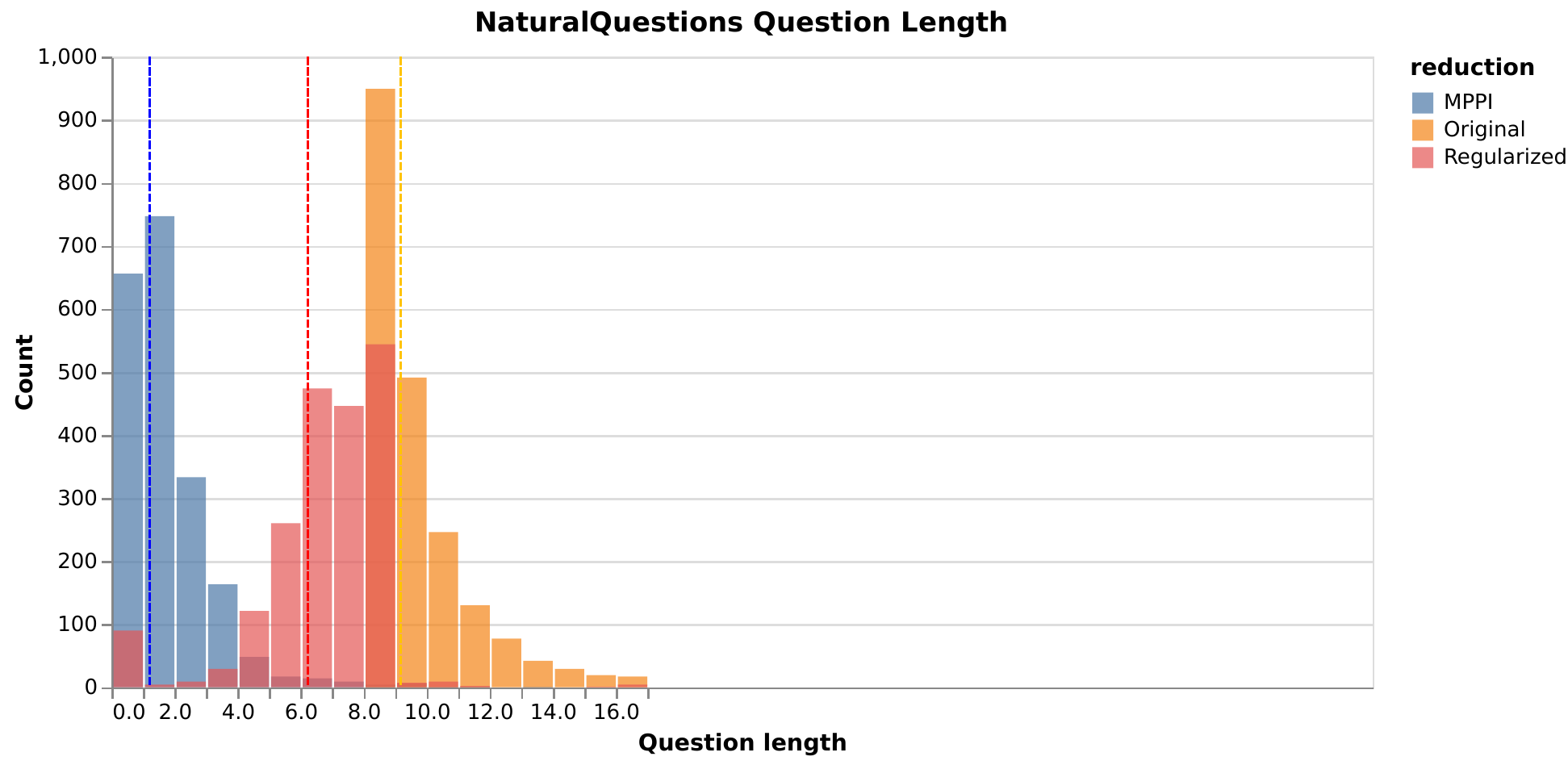}
	\caption{\abr{NatrualQuestion} question length generated by different \mppi{} reduction methods}
	\label{fig:nq-lens}
	\vspace*{-.4cm}
\end{figure}

\begin{figure}[H]
	\includegraphics[width=1\linewidth]{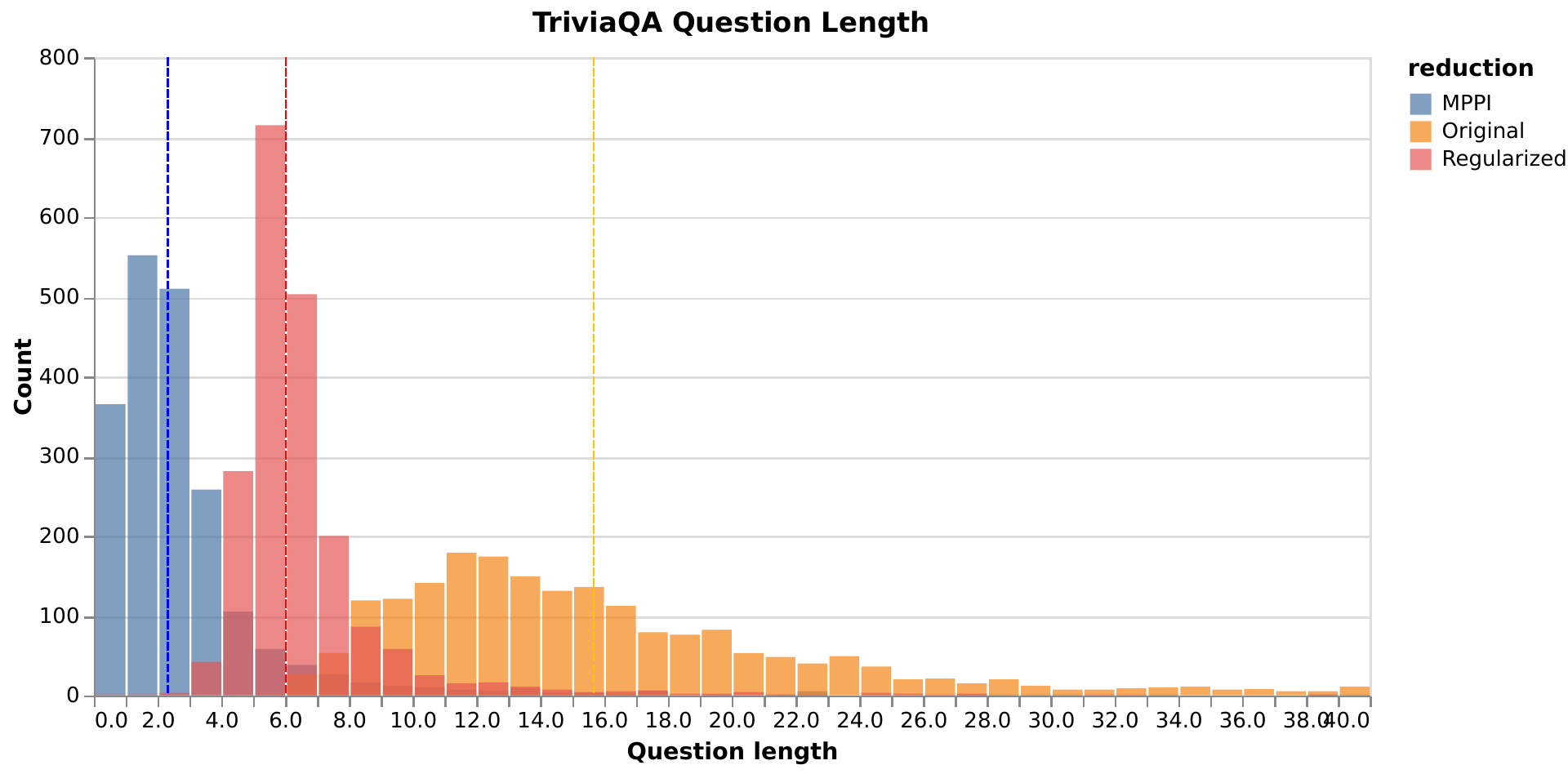}
	\caption{\abr{TriviaQA} question length generated by different \mppi{} reduction methods}
	\label{fig:trivia-lens}
	\vspace*{-.4cm}
\end{figure}

\begin{figure}[H]
	\includegraphics[width=1\linewidth]{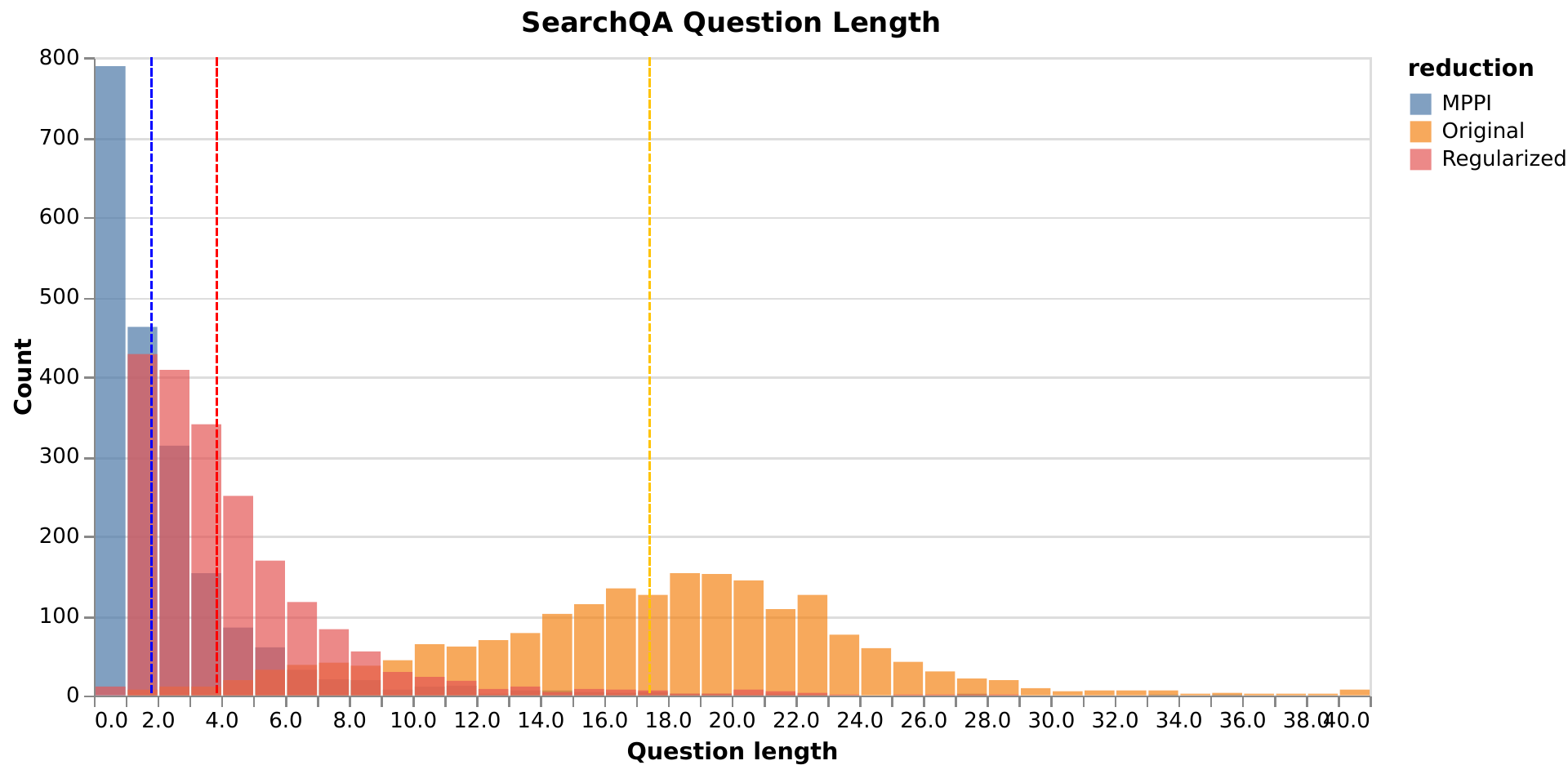}
	\caption{\abr{SearchQA} question length generated by different \mppi{} reduction methods}
	\label{fig:search-lens}
	\vspace*{-.4cm}
\end{figure}

In the main paper we describe the differences in length distributions between original and \mppi{} queries.
To provide more detail into the length distributions we plot histograms of the query word lengths, for the original queries, \mppi{} queries, and \mppi{} queries after the \mppi{} regularization procedure.
These lengths are plotted below for SQuAD (Figure~\ref{fig:squad-lens}), HotpotQA (Figure~\ref{fig:hotpot-lens}), NewsQA (Figure~\ref{fig:news-lens}), Natural Questions (Figure~\ref{fig:nq-lens}), TriviaQA (Figure~\ref{fig:trivia-lens}), and SearchQA (Figure~\ref{fig:search-lens}).


The query length distributions show that \mppi{}s are significantly shorter than original queries, with the \mppi{}s of regularized models somewhere in between.
These length distributions may be sufficient to explain why humans find the non-regularized \mppi{}s completely uninterpretable, and the regularized \mppi{}s somewhat more interpretable.

\section{Are \mppi{}s Invariant to Random Seed?}

One of the preliminary questions in our investigation was whether changing the random training seed significantly altered the \mppi{} produced by a model.
If it were the case that this had a drastic effect, we might suspect \mppi{}s were somewhat random, or the product of meangingless over-confidence on out-of-distribution inputs.
Table~\ref{appendix-random-seed} illustrates the random seed experiment in full. 
Training $10$ SQuAD models, each with different random seeds, we generate \mppi{}s on the $2k$ SQuAD evaluation set, and compare 5 pairs. 
We measure the mean Generalized Jaccard Similarity of \mppi{}s produced by $2$ models trained with different seeds.

We see the similarity between \mppi{}s trained with different seeds far exceed those of Rand-A, and Rand-B, which are akin to a ``random" simulation of \mppi{}s.
As with our previous random baselines these are generated by randomly sampling tokens from the original query, preserving word order, and ensuring that the length distribution matches that of the actual \mppi{}s to which they are being compared.

\begin{table}
	\centering
	\begin{tabular}{llr}
		\hline
		\textbf{Seed A} & \textbf{Seed B} & \textbf{JS / EM} \\
		\hline
		0 & 1 & 55.0 / 31.7 \\
		2 & 3 & 56.8 / 33.2 \\
		4 & 5 & 58.3 / 34.7 \\
		6 & 7 & 57.4 / 33.2 \\
		8 & 9 & 58.1 / 35.2 \\
		\hline
		{}  & Overall & 57.1 / 33.6 \\
		\hline
		Rand-A & Rand-B & 13.8 /  0.9 \\
		\hline
	\end{tabular}
	\caption{\label{appendix-random-seed}
		Observing the Jaccard Similarity and Exact Match between \mppi{}s on the SQuAD $2k$ evaluation set, we see significant token overlap despite seed differences. 
		In contrast, the randomly generated sequences, preserving the length distribution of \mppi{}s, produces far less similar token sequences.
	}
	\vspace*{-.4cm}
\end{table}

\section{Are \mppi{}s Invariant to Training Domain?}

We discussed the invariance of \mppi{}s to training domain at length in the paper for BERT.
For completeness, we provide the raw results for BERT in Table~\ref{bert_dataset-mppi-similarity} and for XLNet in Table~\ref{xlnet_dataset-mppi-similarity}.
These results show that \mppi{}s are far more similar to one another, even when training domain is different.
The random baseline, in parenthesis, once again  shows the Jaccard Similarity we would expect if \mppi{}s were purely random.

\begin{table*}
	\resizebox{\linewidth}{!}{\begin{tabular}{lrrrrrr}
			{\textbf{Train Dataset}} & \multicolumn{6}{c}{\textbf{Reduction Dataset}} \\
			\hline{}
			{} & \abr{SQuAD} & \abr{HotPotQA} & \abr{NewsQA} & \abr{NatrualQ} & \abr{TriviaQA} & \abr{SearchQA} \\
			\hline{}
			\abr{SQuAD} & - (-) & 31.4 (8.8) & 41.0 (21.6) & 29.2 (12.5) & 24.9 (10.9) & 11.9 (9.6) \\
			\abr{HotPotQA} & 39.7 (12.8) & - (-) & 39.6 (18.8) & 33.8 (13.5) & 25.8 (10.7) & 16.6 (12.6) \\
			\abr{NewsQA} & 41.1 (13.0) & 31.6 (7.2) & - (-) & 35.2 (12.5) & 25.8 (10.8) & 13.3 (9.4) \\
			\abr{NatrualQ} & 37.5 (12.7) & 28.7 (7.1) & 40.2 (17.9) & - (-) & 25 (10.7) & 15.0 (11.0) \\
			\abr{TriviaQA} & 33.3 (13.0) & 27.7 (8.0) & 34.8 (18.1) & 29.2 (15.4) & - (-) & 23.1 (15.2) \\
			\abr{SearchQA} & 23.4 (12.3) & 16.8 (7.8) & 24.7 (17.2) & 24.4 (14.6) & 23.4 (11.9) & - (-) \\
			\hline
			{Average} & 35.0 (12.8) & 27.2 (7.5) & 36.1 (18.7) & 30.4 (13.7) & 25.0 (11.0) & 16.0 (11.6) \\
	\end{tabular}}
	\caption{\label{bert_dataset-mppi-similarity}
		The Jaccard Similarity (\%) between \abr{BERT} generated \mppi{}s, across domains. The Random baseline \mppi{}s are in parentheses.
	}
\end{table*}

\begin{table*}
	\resizebox{\linewidth}{!}{\begin{tabular}{lrrrrrr}
			{\textbf{Train Dataset}} & \multicolumn{6}{c}{\textbf{Reduction Dataset}} \\
			\hline{}
			{} & \abr{SQuAD} & \abr{HotPotQA} & \abr{NewsQA} & \abr{NatrualQ} & \abr{TriviaQA} & \abr{SearchQA} \\
			\hline{}
			\abr{SQuAD} & - (-) & 25.8 (9.0) & 37.7 (19.7) & 30.9 (11.1) & 18.1 (10.5) & 22.7 (26.3) \\
			\abr{HotPotQA} & 28.4 (15.3) & - (-) & 31.2 (17.6) & 31.5 (12.4) & 17.8 (12.0) & 27.1 (25.5) \\
			\abr{NewsQA} & 31.8 (13.1) & 25.3 (8.2) & - (-) & 36.6 (11.9) & 20.6 (9.0) & 13.3 (11.7) \\
			\abr{NatrualQ} & 29.9 (12.9) & 24.2 (8.4) & 40.2 (16.8) & - (-) & 22.3 (11.0) & 19.2 (16.3) \\
			\abr{TriviaQA} & 25.6 (14.8) & 19.0 (8.0) & 29.8 (17.4) & 29.2 (13.7) & - (-) & 31.2 (20.6) \\
			\abr{SearchQA} & 21.6 (13.8) & 15.5 (7.7) & 25.2 (15.1) & 24.6 (14.1) & 28.3 (13.4) & - (-) \\
			\hline
			{Average} & 27.5 (14.0) & 22.0 (8.3) & 32.8 (17.3) & 30.6 (12.6) & 21.4 (11.2) & 22.7 (20.1) \\
	\end{tabular}}
	\caption{\label{xlnet_dataset-mppi-similarity}
		The Jaccard Similarity (\%) between \abr{XLNet} generated \mppi{}s, across domains. The Random baseline \mppi{}s are in parentheses.
	}
\end{table*}

\section{Do QA Models Generalize to different \mppi{} Domains?}

\begin{figure}[h]
	\includegraphics[width=1\linewidth]{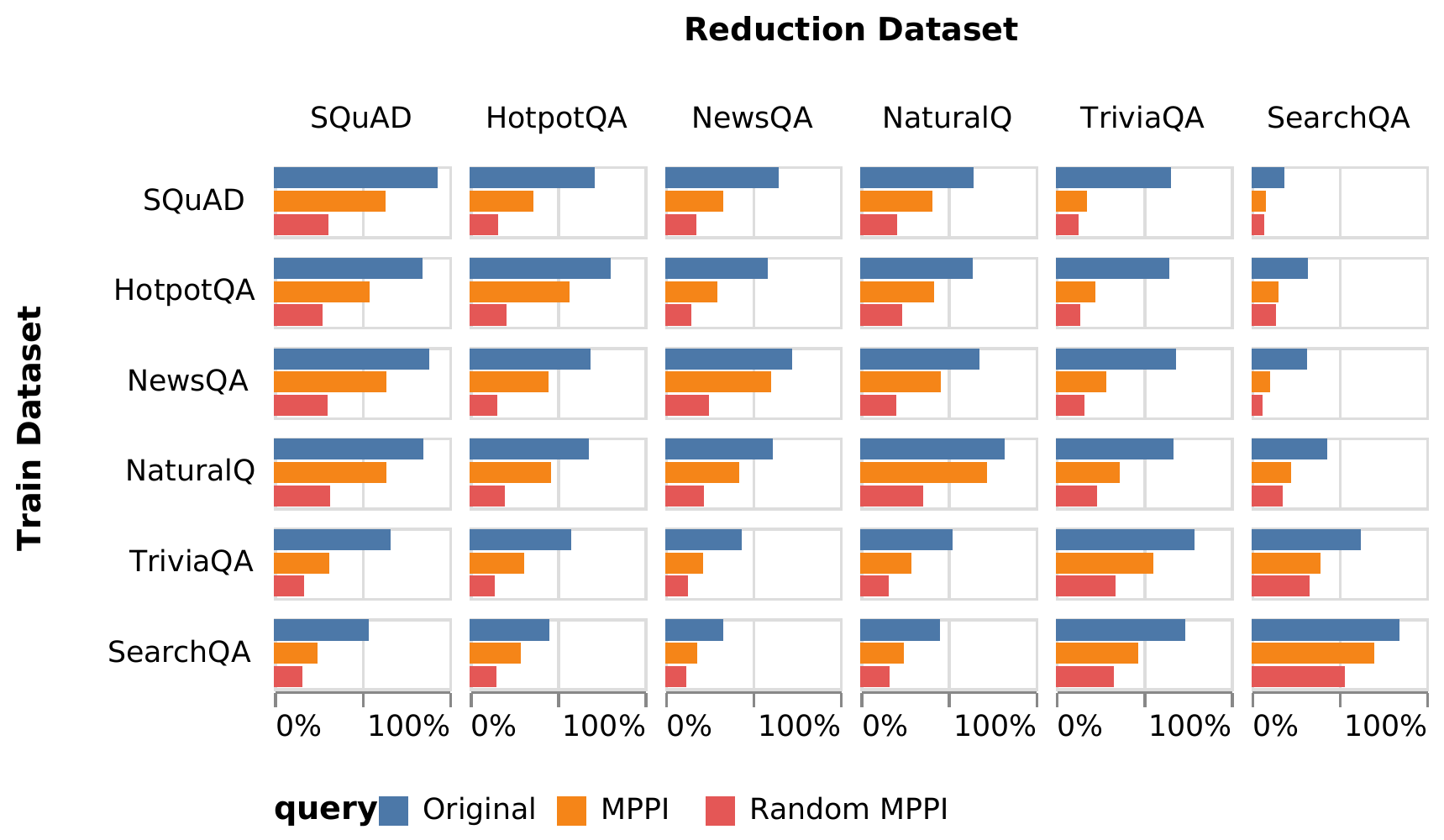}
	\caption{\abr{XlNet} performance with different training sets (y-axis), and evaluation sets (x-axis). Bars measure the F1 score on the $2k$ evaluation set, colored by input type.}
	\label{fig:xlnet-cross-domain-f1}
  \end{figure}

Expanding on the \mppi{} generalization analysis in Section 3.2, we provide the raw results.
The cross-domain generalization of BERT and XLNet models on \mppi{}s sourced from different training domains is available in Table~\ref{appendix-raw-bbc-cross-domain-generalization} and Table~\ref{appendix-raw-xlc-cross-domain-generalization} respectively. Figure~\ref{fig:xlnet-cross-domain-f1} visualizes how well \abr{XLNet} generalizes to different \mppi{} domains.
The results mirror those of \abr{BERT} shown in the main paper.

\begin{table*}
	\small
	\centering
	\begin{tabular}{llrrrrrr}
		\hline
		\textbf{\begin{tabular}{@{}c@{}}Train \\ Dataset\end{tabular}} & \textbf{\begin{tabular}{@{}c@{}}Query \\ Type\end{tabular}} & \textbf{SQuAD} & \textbf{HotpotQA} &\textbf{NewsQA} & \textbf{NaturalQ} & \textbf{TriviaQA} & \textbf{SearchQA}\\
		\hline
		SQuAD & Original & 87.74 & 56.31 & 48.81 & 21.53 & 56.74 & 52.62 \\
		SQuAD & \mppi{} & 87.74 & 28.84 & 31.68 & 13.52 & 43.02 & 30.93 \\
		SQuAD & Random \mppi{} & 26.42 & 16.19 & 19.69 & 9.55 & 13.01 & 17.46 \\
		\hline
		TriviaQA & Original & 54.64 & 71.04 & 42.27 & 47.53 & 51.85 & 34.45 \\
		TriviaQA & \mppi{} & 34.25 & 71.04 & 24.67 & 32.28 & 33.85 & 18.91 \\
		TriviaQA & Random \mppi{} & 15.21 & 32.23 & 15.92 & 25.49 & 14.12 & 12.67 \\
		\hline
		NaturalQ & Original & 75.28 & 58.18 & 77.78 & 37.84 & 54.08 & 51.16 \\
		NaturalQ & \mppi{} & 55.15 & 40.43 & 77.78 & 24.32 & 44.52 & 34.25 \\
		NaturalQ & Random \mppi{} & 23.28 & 23.94 & 38.43 & 18.39 & 16.32 & 17.34 \\
		\hline
		SearchQA & Original & 40.25 & 59.44 & 32.58 & 78.11 & 35.93 & 20.61 \\
		SearchQA & \mppi{} & 24.84 & 41.67 & 21.24 & 78.11 & 24.96 & 15.9 \\
		SearchQA & Random \mppi{} & 11.92 & 24.69 & 14.73 & 45.7 & 12.34 & 9.7 \\
		\hline
		HotpotQA & Original & 71.52 & 53.4 & 52.51 & 38.9 & 75.09 & 47.03 \\
		HotpotQA & \mppi{} & 49.52 & 34.56 & 37.23 & 20.66 & 75.09 & 30.38 \\
		HotpotQA & Random \mppi{} & 21.09 & 19.28 & 24.92 & 15.2 & 17.76 & 17.88 \\
		\hline
		NewsQA & Original & 78.16 & 60.91 & 59.94 & 33.79 & 56.53 & 68.19 \\
		NewsQA & \mppi{} & 61.32 & 39.17 & 42.48 & 19.44 & 48.58 & 68.19 \\
		NewsQA & Random \mppi{} & 22.78 & 20.4 & 22.74 & 15.05 & 14.49 & 24.83 \\
		\hline
	\end{tabular}
	\caption{\label{appendix-raw-bbc-cross-domain-generalization}
		Cross-Domain Generalization of \abr{BERT} Base models on different types of inputs. Values correspond to F1 scores on the question answering $2k$ evaluation set specified by the column.
	}
\end{table*}

\begin{table*}
	\small
	\centering
	\begin{tabular}{llrrrrrr}
		\hline
		\textbf{\begin{tabular}{@{}c@{}}Train \\ Dataset\end{tabular}} & \textbf{\begin{tabular}{@{}c@{}}Query \\ Type\end{tabular}} & \textbf{SQuAD} & \textbf{HotpotQA} &\textbf{NewsQA} & \textbf{NaturalQ} & \textbf{TriviaQA} & \textbf{SearchQA}\\
		\hline
		SQuAD & Original & 93.92 & 64.97 & 66.62 & 15.13 & 70.51 & 65.06 \\
		SQuAD & \mppi{} & 93.95 & 17.75 & 39.95 & 6.37 & 41.36 & 35.27 \\
		SQuAD & Random \mppi{} & 31.0 & 12.86 & 21.12 & 7.3 & 16.19 & 17.85 \\
		\hline
		TriviaQA & Original & 67.55 & 78.15 & 51.7 & 67.69 & 57.85 & 44.97 \\
		TriviaQA & \mppi{} & 31.77 & 78.18 & 27.8 & 43.63 & 32.34 & 22.24 \\
		TriviaQA & Random \mppi{} & 17.01 & 34.03 & 16.03 & 33.18 & 14.54 & 13.27 \\
		\hline
		NaturalQ & Original & 85.61 & 67.84 & 82.06 & 42.92 & 67.43 & 60.82 \\
		NaturalQ & \mppi{} & 63.02 & 35.73 & 82.06 & 20.44 & 46.19 & 42.95 \\
		NaturalQ & Random \mppi{} & 31.74 & 23.31 & 36.07 & 18.05 & 19.98 & 22.45 \\
		\hline
		SearchQA & Original & 55.25 & 74.37 & 45.43 & 84.08 & 45.33 & 32.82 \\
		SearchQA & \mppi{} & 25.6 & 47.5 & 26.33 & 84.08 & 29.41 & 18.57 \\
		SearchQA & Random \mppi{} & 15.92 & 33.29 & 16.81 & 53.58 & 15.57 & 12.13 \\
		\hline
		HotpotQA & Original & 82.85 & 61.03 & 61.93 & 23.98 & 80.28 & 54.19 \\
		HotpotQA & \mppi{} & 51.11 & 14.57 & 40.95 & 9.08 & 80.3 & 23.66 \\
		HotpotQA & Random \mppi{} & 27.83 & 14.0 & 23.89 & 14.03 & 21.33 & 15.25 \\
		\hline
		NewsQA & Original & 88.56 & 69.32 & 67.61 & 30.74 & 69.14 & 73.17 \\
		NewsQA & \mppi{} & 65.64 & 29.66 & 48.56 & 11.0 & 45.15 & 73.12 \\
		NewsQA & Random \mppi{} & 30.47 & 16.29 & 20.27 & 6.64 & 15.71 & 25.35 \\
		\hline
	\end{tabular}
	\caption{\label{appendix-raw-xlc-cross-domain-generalization}
		Cross-Domain Generalization of \abr{XLNet} Large models on different types of inputs. 
		Values correspond to F1 scores on the question answering $2k$ evaluation set specified by the column.
	}
\end{table*}

\end{document}